\documentclass[a4paper]{article}

\usepackage{INTERSPEECH_v2}
\usepackage{latexsym}
\usepackage{graphicx}
\usepackage{url}
\usepackage{tipa}
\usepackage{devanagari}
\usepackage{lipsum}
\usepackage{amsmath}

\title{A Finite State and Rule-based Akshara to Prosodeme (A2P) Converter in Hindi }
\name{Somnath Roy}
\address{
  Centre for Linguistics, Jawaharlal Nehru University, Delhi
  }
\email{somnathroy86@gmail.com}

\begin{document}
\maketitle
\begin{abstract}
         
 This article describes a software module called Akshara to Prosodeme (A2P) converter in Hindi. It converts an input grapheme into prosedeme (sequence of phonemes with the specification of syllable boundaries and prosodic labels). The software is based on two proposed finite state machines\textemdash one for the syllabification and another for the syllable labeling. In addition to that, it also uses a set of nonlinear phonological rules proposed for foot formation in Hindi, which encompass solutions to schwa-deletion in simple, compound, derived and inflected words. The nonlinear phonological rules are based on metrical phonology with the provision of recursive foot structure. A software module is implemented in Python. The testing of the software for syllabification, syllable labeling, schwa deletion and prosodic labeling yield an accuracy of more than 99\% on a lexicon of size 28664 words.
\end{abstract}
\section{Introduction}

The function of Akshara to Prosedeme (A2P) converter is similar to Grapheme to Phoneme (G2P) converter. Moreover, the former also includes the details of syllable boundaries and prosodic labels. Past studies on Hindi has mainly focused on the implementation of G2P converter for speech synthesis programme \cite{bali2004tools,narasimhan2004schwa,choudhury2003rule}. These G2P converters are based on linear phonological rules proposed by \cite{ohala1983aspects}. However, syllable is known to be a better unit for Hindi speech synthesis as reported in \cite{bellur2011prosody,kishore2003unit}. The usefulness of syllable as the basic linguistic unit in the context of speech recognition system has been explored in English \cite{ganapathiraju2001syllable} and Tamil \cite{lakshmi2006syllable}. Similar work for Hindi requires a software for syllabification. The automatic syllabification would be more useful if it could also predict the prosodic labels in words of natural speech as this would facilitate synthesis and recognition. This work fulfills that need. The A2P converter is based on the following novel contributions.

\begin{itemize}

\item A finite state machine (FST) is proposed for both syllabification and syllable labeling. The use of FSTs yield faultless syllabification and syllable labeling at underlying phonemic form. The syllabification at underlying phonemic form is called as I-level syllabification in this work. 

\item A new foot formation rule is proposed. The proposed rules assume the extrametricality of foot unlike syllable as proposed in \cite{pandey2014akshara} and allow superfoot by prosodic recursion described in \cite{ito2012recursive,martinez2012superfeet}. 

\item The proposed foot formation rules are autonomous to morphological boundaries. The rule set performs better than the previous works for schwa deletion in simple, compound, derived and inflected words.

\end{itemize}

Rest of this paper is organized as follows. Section 2 describes the salient points of metrical phonology relevant to this work. Section 3 describes the  process  of  syllabification  and  syllable labeling. Section 4 describes the proposed foot formation rule. Section 5 describes schwa deletion and prosodic labeling. Section 6 describes the testing and evaluation report. The conclusion is written in Section 7. 

\section{Theoretical Background}

Metrical phonology is based on nonlinear arrangement of the constituents of a phrase  \cite{hayes1995metrical,liberman1977stress,selkirk1980role,hayes1980metrical,selkirk1986phonology,apoussidou2006learnability}. The nonlinear arrangement is realized in the form of a tree with nodes as the constituents of a phrase. The constituents are syllable, foot, phonological word, phonological phrase and intonational phrase. Syllable is the lowest unit in the hierarchy dominated by foot, which in turn is dominated by a phonological word. The higher units such as phonological phrase and intonational phrase are not relevant in the current work.
Syllable functions as a domain for segmental phonological rules. In non-linear phonology, the rules are written on the basis of interaction among syllables under the domain of higher constituents. A syllable has obligatory rhyme and optional coda. The syllables are also described by the moraic weight in quantity-sensitive languages such as Hindi  \cite{pandey1989word}. 
Foot as a domain is used for describing stress and re-syllabification due to deletion of segment like schwa in languages such as French and Hindi. The foot is used as a musical meter and the concept is borrowed to non-linear phonology as a constituent  \cite{selkirk1980role,hayes1995metrical}. Most of the quantity sensitive languages have binary foot, but some also allow degenerate foot. A binary foot is erected on either two syllables or on a single syllable having at least two moras. A single syllable having one mora, if projected as a foot, is called degenerate foot  \cite{liberman1977stress,hayes1995metrical}.
Phonological word, also known as prosodic word, is a constituent unit of prosodic hierarchy above syllables or foot and below phonological phrases. Prosodic word takes into account the non-isomorphism between morphology and phonology \cite{hall1999studies}. In other words, a morphological word may not be a phonological word. Prosodic word also accommodate derivational affixes and compounding, and characterized as the domain for syllabification and stress rules.  

\section{I-Level Syllabification}

I-level syllabification is derived from the underlying phonemic form (UPF), which in turn is derived from orthography using the following mapping rules.
\begin{itemize}

\item Each consonant grapheme in Hindi is inherently associated with the mid-central vowel called schwa ("\textipa{@}").
\item If a consonant is followed by a vowel diacritic mark, or a diacritic called halant, the inherent schwa is deleted.
\item The inherent schwa is not realized in case of consonant at word final position.
\item Two or three consonant together can form a ligature. 
\item A short vowel at word final position is lengthened. 
\end{itemize}
For example, {\dn kml} /k m l/ (Lotus) becomes {\textipa{k@m@l}, {\dn kmAl} /k m A l/ (Wonder) becomes {\textipa{k@ma:l}}, {\dn s(y} /s t j/ (Truth) becomes {\textipa{s{@}tj}}, {\dn aEt} /a t i/ (Excess) becomes {\textipa{@ti:}}

The process of syllabification in Hindi was explored by \cite{ohala1983aspects} and \cite{pandey1989word,pandey2014akshara}. Their analysis do not talk about the role of maximal onset principle for syllabification. The present analysis for syllabification follows maximum onset principle \cite{selkirk1984major}. The maximum onset principle is a sufficiency condition which can be demonstrated by the following examples.
\begin{itemize}

\item {\dn \9m(y\7\2jy} (A name) $\rightarrow$ {[mri] [tjun] [d\textipa{Z@}j]} 

\hspace{50pt}$\rightarrow$ *{[mrit][jun][d\textipa{Z@}j]}
                                             
\hspace{50pt}$\rightarrow$ *{[mrit][jund\textipa{Z}][\textipa{@}j]}

\item  {\dn aDEg\5hZ} (Capture) $\rightarrow$ {[\textipa{@}] [d\super hi] [gr\textipa{@}] [h\textipa{@}n]} 

\hspace{50pt}$\rightarrow$ *{[\textipa{@}] [d\super hig] [r\textipa{@}] [h\textipa{@}n]}

\hspace{50pt}$\rightarrow$ *{[\textipa{@}d\super h] [ig] [r\textipa{@}] [h\textipa{@}n]}

\end{itemize}

In the above examples, the right hand side shows the potential syllable structures for a word. The square bracket denotes the syllable boundary.  An asterisk before the syllable structure indicates that this potential syllable sequence is incorrect. Moreover, the above examples show that either onsets are maximized or is equal to number of coda consonants. Therefore, maximum onset principle need to be applied to ensure correct syllabification. Based on many such examples, following regular expressions are proposed for syllabification in Hindi.
\begin{itemize}
\item  v $\rightarrow$ [v]
\item  vv $\rightarrow$ [v][v]
\item  c\super *vcv $\rightarrow$ [c\super * v] [cv]
\item  c\super *vc1cv $\rightarrow$ [c\super *vc1] [cv]
\item  c\super *vc1c2v $\rightarrow$ [c\super *v] [c1c2v]
\item  c\super *vc1c1v $\rightarrow$ [c\super *vc1] [c1v]
\item  c\super *vc2c2v $\rightarrow$ [c\super *vc2] [c2v]
\item  c\super *vcccv $\rightarrow$ [c\super *vc] [ccv]
\end{itemize}

In the above expressions, v denotes a vowel, "c1" denotes a stop consonant, "c2" represents a semivowel (r, l, \textipa{v}, j) and "c" at intervocalic position denotes a consonant that is neither a stop nor a semivowel. An asterisk in superscript denotes the kleene star.

The finite state machine for the I-level syllabification is shown in Fig 1. It contains seventeen states with the start state as I and the  final state as F. The orthography of an input word is transliterated into a sequence of consonants and vowels. The 8 syllabification rules are applied to this sequence to derive a symbol sequence in terms of c, c1, c2 and v. The symbol sequence is the input to the FSM in Fig 1. An arc between a pair of states in FSM is associated with a label (a pair of symbols separated by "/"). Suppose the symbol pair associated with an arc is "x/y". This indicates that whenever a symbol "x" is fed to the state at the beginning of the arc, the system makes a transition along the arc and outputs the symbol "y". The label e/e symbolizes null input and null output for a transition. If part of a symbol string reaches to the final state F, then it is consumed, and a transition from F to I with arc label e/b takes place, where e is null and b denotes the syllable boundary of the consumed string. The remaining part of the string repeats the same process from initial state I until everything is consumed. \\ 
\begin{figure}[h]
\centering
\includegraphics[scale=0.27]{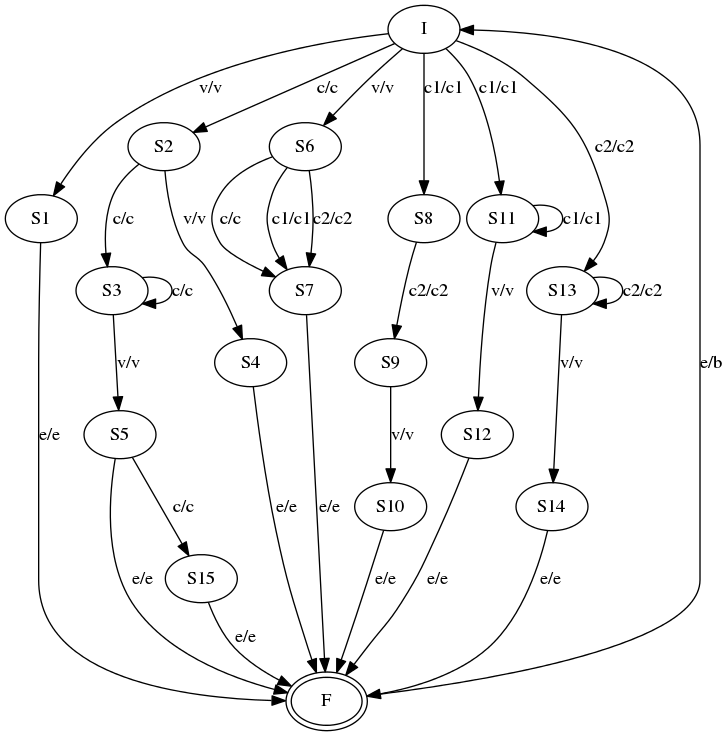}
\caption{Finite State Machine for Syllabification}
\end{figure}

\subsection{Syllable Labeling}
Hindi is a quantity sensitive language. Therefore, syllables in Hindi are also described based on an attribute called moraic weight \cite{hayes1980metrical,pandey1989word}. The stress assignment rules for English proposed by \cite{church1985stress} reports moraic weight (syllable weight) as the most influential parameter. Based on moraic weight, the following rules are used for syllable labeling in Hindi.

i. Each short vowel (like \textipa{@,u,i} ) and each coda consonant of a syllable are assigned a weight of one mora, while a long vowel (like \textipa{a:,u:,i:}) is assigned a weight of two moras.

ii. The syllables with one, two and three moras are called weak (w), heavy (h) and superheavy (sh) syllables respectively \cite{pandey1989word,hayes1989compensatory,pandey1990hindi,kachru2006hindi}.

The examples of syllable labels are shown below in the Table 1 and Table 2. The Table 1 shows monosyllabic words and Table 2 shows bisyllabic words with syllable boundary denoted by square bracket and syllable labels in terms of w, h, and sh.  
\begin{table}[h]
\begin{center}
\begin{tabular}{|l|l|l|l|}

\hline
{ Word} & { Type} & { Syllable Label} & { Gloss}\\\hline
{ki} & {w}  & {[ki]\super w} & {that}\\\hline
{m\textipa{@}n} & {h}  & {[m\textipa{@}n]\super h} & {soul}\\\hline
{g\textipa{a:}l} & {sh}  & {[g\textipa{a:}l]\super {sh}} & {cheek}\\\hline
\end{tabular} 
\caption{\label{Monosyllabic words with syllable labels} Monosyllabic words with syllable labels}
\end{center}
\end{table}

\begin{table}[h]
\begin{center}
\begin{tabular}{|l|l|l|l|}

\hline
{ Word} & { Type} & { Syllable Label} & { Gloss}\\\hline
{k\textipa{@}l\textipa{@}m} & {w+h}  & {[k\textipa{@}]\super w[l\textipa{@}m]\super h} & {pen}\\\hline
{r\textipa{e:}s\textipa{@}m} & {h+h}  & {[r\textipa{e:}]\super h [s\textipa{@}m]\super h} & {silk}\\\hline
{\textipa{a:}r\textipa{a:}m} & {h+sh}  & {[\textipa{a:}]\super h[r\textipa{a:}m]\super {sh}} & {comfort}\\\hline
{d\textipa{e:v}n\textipa{a:}r} & {sh+sh}  & {[d\textipa{e:v}]\super {sh}[n\textipa{a:}r]\super {sh}} & {A name}\\\hline

\end{tabular} 
\caption{\label{Bisyllabic words with syllable labels} Bisyllabic words with syllable labels}
\end{center}
\end{table}

A  finite  state  machine  for  syllable  labeling  is
shown in Fig 2.  The machine consists
of one initial state (state I),  seven non-final states and three
final states (F1, F2 and F3).  The syllabified string (i.e, the syllable boundary marked as b) is given as input to the initial state. Since, each short vowel gets one mora and long vowels get two moras, therefore, vowel type distinction is essential at syllable labeling stage. Each coda consonants get one mora and the onset consonants do not contribute to the moraic weight of syllables in Hindi. Therefore, consonant type distinction is not required at this stage. The symbol c, v\_s and v\_l stand for  consonants, short vowels and  long vowels respectively. The output symbol along an arc is either a syllable label or the reflection of the input itself. There is a null transition from each final state to the initial state so that the process can be repeated for the remaining part of the string. The FSM assigns the label "w" to a syllable with phoneme sequences v\_s, cv\_s, ccv\_s, ccc\super *v\_s and "h" to v\_sc , c\super *v\_sc, c\super *v\_l, and "sh" to c\super *v\_scc, c\super *v\_lc.\\

\begin{figure}[h]
\centering
\includegraphics[scale=0.3]{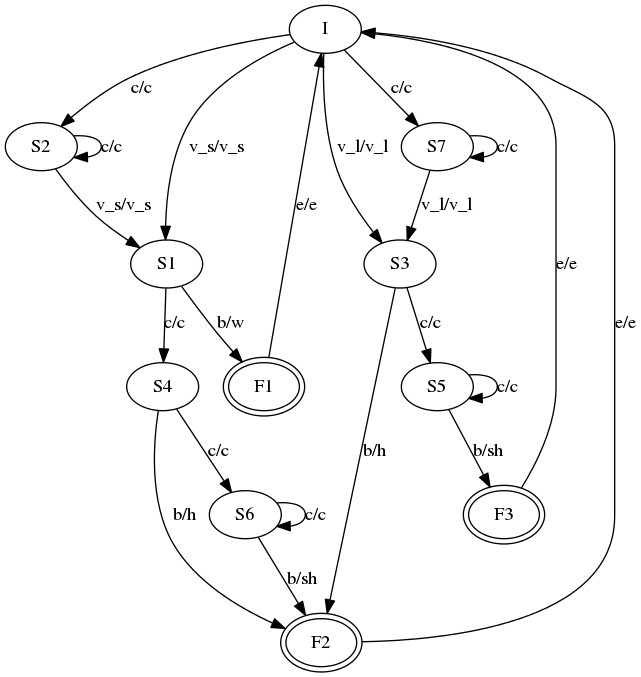}
\caption{Finite State Machine for Syllable Labeling}
\label{Figure1}
\end{figure}

\section{Foot Formation}
This paper describes a new approach to foot formation in Hindi. The approach is based on three assumptions, six rules and one convention as described below.

\subsection{Assumptions}
\begin{itemize}

\item Foot is formed using the labeled syllables of a word. The process of syllable labeling is described in the section 3.
\item Foot is generally binary branching. A bimoraic foot and degenerate foot projected on a single syllable are also permissible.
\item A superfoot is formed either between a syllable and a foot, or between two foot.
\end{itemize}

\subsection{Rules}
The six rules for foot formation are listed below.
\begin{itemize}

    \item Weak to Weak Affinity Rule (WWAR): Two adjacent weak syllables form a binary foot and results into a bimoraic foot as shown in Fig 3. The foot \textless$\sum_{s}$\textgreater is the extra metrical foot, which never bears any stress.\\
\begin{figure}[h]
\centering
\includegraphics[scale=0.4]{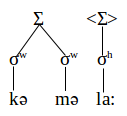}
\qquad
\includegraphics[scale=0.4]{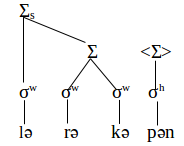}
\caption{Bimoraic binary foot}
\label{Figure1}
\end{figure}

\item Heavy to Weak Affinity Rule (HWAR): A heavy and a weak adjacent syllables form trimoraic binary foot as shown in Fig 4.\\

\begin{figure}[h]
\centering
\includegraphics[scale=0.4]{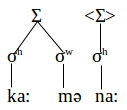}
\qquad
\includegraphics[scale=0.4]{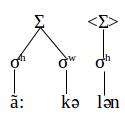}
\caption{Trimoraic foot}
\label{Figure1}
\end{figure}

\item Weak to Heavy Affinity Rule (WHAR): This kind of foot is either formed in bisyllabic Hindi words or in loan words as shown in Fig 5. \\
\begin{figure}[h]
\centering
\includegraphics[scale=0.4]{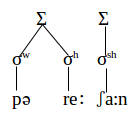}
\qquad
\includegraphics[scale=0.4]{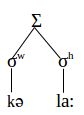}
\caption{Trimoraic foot}
\label{Figure1}
\end{figure}

\item Heavy to Heavy Affinity Rule (HHAR): Two adjacent heavy syllables also form a binary foot. In this case trisyllabic words having left most syllable as heavy is downgraded as weak and bisyllabic words having right most node as heavy is downgraded as weak. The foot structure is shown in Fig 6 \\
\begin{figure}[h]
\centering
\includegraphics[scale=0.4]{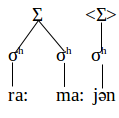}
\qquad
\includegraphics[scale=0.4]{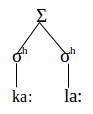}
\caption{Heavy syllables forming a foot}
\label{Figure1}
\end{figure}

\item Superheavy to Others Affinity Rule (SOAR): The superheavy syllables always projected as a foot as shown below in Fig 7. The superfoot ($\sum_{s}$) is formed using one syllable and one foot. The projected foot constituent in $\sum_{s}$ could be either a new syllable after schwa deletion as shown in Fig 3 or a syllable which inherently bear stress i.e., the superheavy syllable as shown in Fig 7.\\

\begin{figure}[!h]
\centering
\includegraphics[scale=0.4]{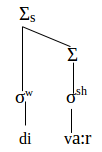}
\caption{Superheavy syllables forming foot}
\label{Figure1}
\end{figure}

\item List Affinity Rule (LAR): LAR is devised for handling words having same syllable structure at underlying phonemic form but realized differently at surface level. In such overlapping cases different foot formation rules are applied. These rules make categorical distinction based on the segmental information in a syllable. The foot formation rules shown in Fig. 8 describes four cases with overlapping syllable structure. These four cases represent four different lists of words.\\
\begin{figure}[h]
\centering
\includegraphics[scale=0.4]{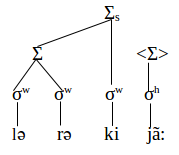}
\qquad
\includegraphics[scale=0.4]{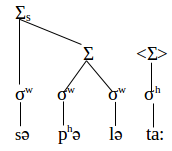}
\qquad
\includegraphics[scale=0.4]{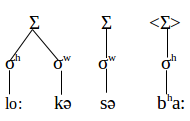}
\qquad
\includegraphics[scale=0.4]{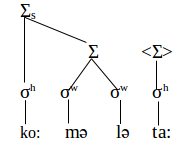}
\caption{Examples of words with same syllable structure but different foot formation}
\label{Figure1}
\end{figure}

\end{itemize}

\subsection{ Convention}
The rules apply in the direction from left to right (LR).

These rules are called as affinity hierarchy rules. The word affinity describes the interaction between different or similar type of syllables. The word 'hierarchy' is used because these rules apply in the order of their height. The top rule in the hierarchy applies first and so on. The decreasing order of height of these rules is LAR\textgreater WWAR \textgreater HWAR \textgreater WHAR \textgreater HHAR \textgreater SOAR

\section{Schwa Deletion and Prosodic Labeling}

Schwa deletion  is an optional phenomenon in Hindi. This means that schwa can be deleted or retained in the same environment, and the choice solely depends on the speaker and the context being used. Schwa deletion phenomena in Hindi helps speakers to utter a word quickly, i.e., the process of schwa deletion reduces the overall effort in terms of duration. It enables stress shift from one syllable to other. The following rules describe the contexts in which schwa gets deleted.
\begin{itemize}
    \item \textipa{@} $\rightarrow \Phi$ / {[\bf{$\sigma$}\super w \textemdash \bf{$\sigma$}\super w ]}$\sum$
    \item \textipa{@} $\rightarrow \Phi$ / [\bf{$\sigma$}\super h \textemdash \bf{$\sigma$}\super w ]$\sum$ 
    \item  \textipa{@} $\rightarrow \Phi$ / [\bf{$\sigma$}\super {sh} \textemdash \bf{$\sigma$}\super w ]$\sum$ 
\end{itemize}

 In other words, if the right most node of a foot is a weak syllable having schwa then schwa can be deleted.

Application of the above schwa deletion (SD) rules and consequent re-syllabification of exemplar words are shown in Table 3.
The process of schwa deletion and re-syllabification occur at foot level. In the process of re-syllabification, the bare consonant(s) after schwa deletion are assigned as coda consonant(s) to the preceding syllable. The re-syllabified words are again fed to the finite state machine for syllable labeling. The syllable label associated to a syllable after re-syllabification is called the prosodic label. If prosodic label of a syllable is heavy or superheavy then it is stressed otherwise not stressed. The foot structure for the examples in Table 3 can be found in the Section 4. 

\begin{table}[h]
\addtolength{\tabcolsep}{-5pt}
\begin{center}

\begin{tabular}{|l|l|l|l|}
\hline
 { UPF} &  { SD} & { Resyllab}& { Gloss}\\
\hline
{k\textipa{@}m\textipa{@}l\textipa{a:}} & {k\textipa{@}ml\textipa{a:}} &{[k\textipa{@}m][l\textipa{a:}]}  & {A Name}\\\hline
{l\textipa{@}r\textipa{@}k\textipa{@}p\textipa{@}n} & {l\textipa{@}r\textipa{@}kp\textipa{@}n} &{[l\textipa{@}][r\textipa{@}k][p\textipa{@}n]}  & {Childhood}\\\hline
{k\textipa{a:}m\textipa{@}n\textipa{a:}} & {k\textipa{a:}mn\textipa{a:}} &{[k\textipa{a:}m] [n\textipa{a:}]}  & {Wish}\\\hline
{l\textipa{o:}k\textipa{@}s\textipa{@}b\super h\textipa{a:}} & {l\textipa{o:}ks\textipa{@}b\super h\textipa{a:}} &{[l\textipa{o:}k][s\textipa{@}][b\super h\textipa{a:}]}  & {Parliament}\\\hline
{s\textipa{@}p\super h\textipa{@}l\textipa{@}t\textipa{a:}}&{s\textipa{@}p\super h\textipa{@}lt\textipa{a:}} &{[s\textipa{@}] [p\super h\textipa{@}l] [t\textipa{a:}]}  & {Success}\\\hline
\end{tabular} \\
\end{center}
\caption{\label{Examples of word re-syllabification (Resyllab) after applying schwa deletion (SD) to underlying phonemic form (UPF)} Examples of word re-syllabification (Resyllab) after applying schwa deletion (SD) to underlying phonemic form (UPF)}
\end{table}

\section{Testing and Evaluation}
Two databases are used for the testing and evaluation \textemdash BBC Hindi corpus and Hindi Wordnet \cite{bhattacharyya2010indowordnet}. Hindi wordnet is publicly accessible. It is first cleaned i.e., digits, hyphen and other special characters are removed. Long but meaningful words especially compounds, derived and inflected words are picked up from different lexical categories like Noun, Verb, Adjective and Adverbs. Finally, a list of 5817 words is collected in which schwa deletion rule applies at least once.  The databases used for the testing and evaluation by the previous systems like  \cite{bali2004tools},\cite{narasimhan2004schwa}, \cite{choudhury2003rule} and \cite{raj2007text} are not publicly available. Therefore, an another word list of size 22847 words is also used for the testing and evaluation. These words are extracted from the BBC Hindi corpus and it contains the same 16000 words used in \cite{pandey2014akshara}. The word lists from both the databases are fed to the software and resultant output is verified against the list annotated by an expert at four levels; a) Syllabification  b) Syllable Labeling c) Schwa deletion and d) Stress Alignment using Prosodic Label. The summary of test result is shown below in Table 4.
\begin{table}[h]
\addtolength{\tabcolsep}{-5pt}
\begin{center}

\begin{tabular}{|l|l|l|l|}
\hline
{Testing Level} & {\#Words}& { \#WO}& { \% Accuracy}\\\hline
{Syllabification} & {28664}  & {Nil} &{100}\\\hline
{Syllable Labeling} & {28664}  & {Nil} &{100}\\\hline
{Schwa Deletion} & {28664}  & {121} &{99.58}\\\hline
{Stress Alignment using Prosodic Label} & {28664}  & {139} &{99.52}\\\hline
 
\end{tabular} 
\end{center}
\caption{\label{A summary of testing of the current implementation for syllabification, Syllable Labeling, Schwa Deletion and Prosodic Labeling}A summary of testing of the current implementation for syllabification, Syllable Labeling, Schwa Deletion and Prosodic Labeling}
\end{table}

\section{Conclusion}

 In this paper, a software module A2P converter is described. The software is based on two finite state machines and non-linear phonological rules. The A2P converter shows an accuracy of more than 99\% at all the testing level. The current implementation cannot be compared with existing systems because the past systems are only for G2P conversion. However, the A2P converter shows an improvement of more than 2\% for G2P conversion over the best existing G2P converters.

\bibliographystyle{IEEEtran}

\bibliography{mybib}

% \begin{thebibliography}{9}
% \bibitem[1]{Davis80-COP}
%   S.\ B.\ Davis and P.\ Mermelstein,
%   ``Comparison of parametric representation for monosyllabic word recognition in continuously spoken sentences,''
%   \textit{IEEE Transactions on Acoustics, Speech and Signal Processing}, vol.~28, no.~4, pp.~357--366, 1980.
% \bibitem[2]{Rabiner89-ATO}
%   L.\ R.\ Rabiner,
%   ``A tutorial on hidden Markov models and selected applications in speech recognition,''
%   \textit{Proceedings of the IEEE}, vol.~77, no.~2, pp.~257-286, 1989.
% \bibitem[3]{Hastie09-TEO}
%   T.\ Hastie, R.\ Tibshirani, and J.\ Friedman,
%   \textit{The Elements of Statistical Learning -- Data Mining, Inference, and Prediction}.
%   New York: Springer, 2009.
% \bibitem[4]{YourName17-XXX}
%   F.\ Lastname1, F.\ Lastname2, and F.\ Lastname3,
%   ``Title of your INTERSPEECH 2017 publication,''
%   in \textit{Interspeech 2017 -- 18\textsuperscript{th} Annual Conference of the International Speech Communication Association, August 20?24, Stockholm, Sweden, Proceedings, Proceedings}, 2017, pp.~100--104.
% \end{thebibliography}

\end{document}